\documentclass[sigconf]{acmart}

\usepackage{multirow}
\usepackage{graphicx}
\usepackage{subfig}
\usepackage{algorithm,algorithmic}
\usepackage{bm}
\AtBeginDocument{%
  }

\copyrightyear{2026}
\acmYear{2026}
\setcopyright{cc}
\setcctype{by}
\acmConference[KDD '26]{Proceedings of the 32nd ACM SIGKDD Conference on Knowledge Discovery and Data Mining V.2}{August 09--13, 2026}{Jeju Island, Republic of Korea}
\acmBooktitle{Proceedings of the 32nd ACM SIGKDD Conference on Knowledge Discovery and Data Mining V.2 (KDD '26), August 09--13, 2026, Jeju Island, Republic of Korea}
\acmDOI{10.1145/3770855.3818397}
\acmISBN{979-8-4007-2259-2/2026/08}

\begin{document}


\title{Empowering Credit Risk Detection in Weixin Pay with Billion-Scale Deep Graph Learning}

\author{Xin Liu}
\email{2432042@tongji.edu.cn}
\orcid{0009-0007-8007-1560}
\affiliation{%
  \institution{Tongji University}
  \city{Shanghai}
  \country{China}
}

\author{Xiyuan Chen}
\email{xiyuanchen@tencent.com}
\orcid{0009-0008-8408-338X}
\affiliation{%
  \institution{Tencent Inc.}
  \city{Shenzhen}
  \country{China}}

\author{Chenglong Wu}
\email{leonclwu@tencent.com}
\orcid{0009-0004-9096-3843}
\affiliation{%
  \institution{Tencent Inc.}
  \city{Shenzhen}
  \country{China}
}

\author{Xuan Zong}
\email{jasonzong@tencent.com}
\orcid{0009-0003-4109-8438}
\affiliation{%
  \institution{Tencent Inc.}
  \city{Shenzhen}
  \country{China}
}

\author{Jun Zhou}
\email{anderszhou@tencent.com}
\orcid{0009-0008-8182-4795}
\affiliation{%
  \institution{Tencent Inc.}
  \city{Shenzhen}
  \country{China}
}

\author{Dawei Cheng}
\authornote{Corresponding Author.}
\email{dcheng@tongji.edu.cn}
\orcid{0000-0002-5877-7387}
\affiliation{%
  \institution{Tongji University}
  \city{Shanghai}
  \country{China}
}

\renewcommand{\shortauthors}{Xin Liu et al.}

\begin{abstract}
Credit risk detection, particularly mitigating individual fraud, is crucial for maintaining the stability of digital financial ecosystems. Accurately identifying credit fraud among billions of users is critical for minimizing financial losses and safeguarding the sustainability of inclusive financial services. 
Given that credit fraud risks are often concealed within heterogeneous user-risk graphs, Graph Neural Networks (GNNs) have emerged as an effective tool for risk mining by capturing complex dependencies. To address the scalability bottleneck of industrial GNNs, distributed training based on subgraphs is indispensable. However, existing strategies often compromise topological integrity for load balancing. This can be catastrophic for risk detection, as it indiscriminately severs the long-tail evidence chains essential for risk propagation. Overlapping subgraphs can restore severed risk contexts but inevitably introduce redundancy and noise, while overlooking the representation alignment across different local subgraphs.
In this paper, we propose a risk-aware overlapping subgraph learning framework for large-scale credit risk detection.
We first construct base partitions to ensure load balance. Then, we perform budget-constrained sampling that selects informative long-tail nodes, thereby preserving critical risk diffusion patterns while filtering out noise. To mitigate representation inconsistency, we design a cross-subgraph consistency alignment mechanism. By enforcing alignment constraints on the overlapping nodes, we harmonize the local representations into a globally consistent latent space.
Extensive experiments on Weixin Pay's production dataset demonstrate that our model significantly outperforms existing strategies for risk detection, offering a scalable and effective solution for industrial graph learning.
\end{abstract}

\ccsdesc[500]{Computing methodologies~Machine learning}
\ccsdesc[500]{Applied computing}
\ccsdesc[500]{Information systems~Data mining}


\keywords{Credit Risk Detection, Graph Fraud Detection, Large-scale Graph}


\maketitle

\section{Introduction}
Credit serves as a fundamental engine for economic growth by facilitating the intertemporal allocation of resources~\cite{bernanke1988credit, bernanke1993credit}. Fraudsters maliciously obtaining credit facilities not only causes direct losses to financial institutions but also threatens overall financial stability, especially when credit demand exceeds available financial resources~\cite{stulz2010credit, brown2014credit, caouette1998managing}. Consequently, financial institutions and enterprises must adopt precise risk assessment methodologies~\cite{crouhy2000comparative, chen2016financial}. Effectively allocating credit while rigorously mitigating fraud risks can safeguard the resilience and economic sustainability of the entire financial infrastructure. 

\begin{figure}
    \centering
    \includegraphics[width=1\linewidth]{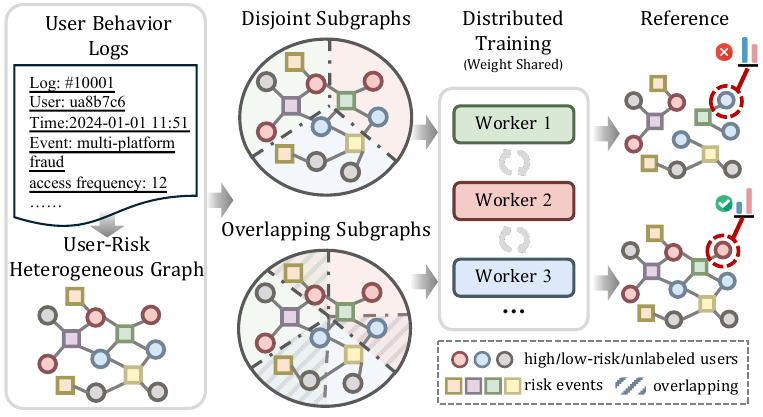}
    \caption{Comparison of subgraph-based strategies. User behaviors are modeled as a heterogeneous graph. Unlike disjoint partitioning, the overlapping subgraphs preserve critical risk diffusion paths to guarantee accurate risk prediction.}
    \label{fig:comparison}
\end{figure}

Existing risk detection has relied on expert rules and machine learning models driven by statistical feature engineering~\cite{seeja2014fraudminer,fu2016credit,niu2018visual}. These approaches fail to capture non-linear characteristics and adapt to evolving risk patterns. Effective risk detection must transcend the evaluation of isolated user profiles, which requires mining the intrinsic correlations between user behaviors and risk events to uncover latent high-risk users. Recently, Graph Neural Networks (GNNs) have emerged as a dominant paradigm in financial fraud detection owing to their ability to capture potential dependencies~\cite{cheng2025graph, xiang2023semi, wu2025defending}. As shown in Figure~\ref{fig:comparison}, a heterogeneous graph comprising nodes of user and risk events can depict the credit risk landscape~\cite{xiang2022temporal}. For instance, a risk event such as an adverse credit record is modeled as a specific risk node connected to the users involved. Through the modeling paradigm, explicit semantic information is propagated to user nodes via a message-passing mechanism. This facilitates end-to-end learning of risk attribution and complex propagation patterns, allowing the model to identify the sources and pathways of potential fraud risk. 

Despite their effectiveness, deploying GNNs in an industrial setting faces a severe scalability bottleneck. Real-world graphs typically scale to billions of nodes and edges. To address this efficiency challenge, extensive research has been conducted~\cite{lin2023comprehensive, wu2021seastar, zhou2023ugrapher}. One branch focuses on decoupling feature propagation and transformation~\cite{wu2019simplifying, chen2020scalable}. For example, HopGNN~\cite{chen2023node} simplifies GNNs into MLP-like architectures via pre-computed propagation. NodeFormer~\cite{wu2022nodeformer} utilizes linearized attention to bypass recursive message passing. However, these methods rely on fixed propagation schemes that lack the adaptive sensitivity required for risk detection. Consequently, sampling-based strategies~\cite{zeng2021decoupling} remain the dominant paradigm for industrial applications due to their compatibility with inductive learning. While node-wise methods like GraphSAGE~\cite{hamilton2017inductive} control the receptive field by sampling fixed neighbors, they suffer from the neighbor explosion problem. To mitigate this, subgraph-based methods~\cite{chiang2019cluster,zeng2020graphsaint} and learnable pooling strategies~\cite{gao2018large} decompose the graph into manageable subgraphs to enable parallel training. 
Some subgraph-based strategies prioritize computational load balance by partitioning the graph into disjoint subgraphs. As Figure~\ref{fig:comparison} shows, these methods can be catastrophic for risk patterns characterized by subtle long-tail dependencies. They indiscriminately break vulnerable evidence chains essential for risk propagation, leading to suboptimal detection accuracy.
In this context, overlapping subgraphs can be beneficial, enabling distinct subgraphs to share risk contexts via overlapping nodes. Nevertheless, effective selection is critical: (1) how to selectively preserve informative risk patterns for keeping detection accuracy while filtering out redundancy and noise to ensure efficiency; (2) how to maintain representation consistency across distributed subgraphs.

To bridge the gap between industrial scalability and high-precision risk detection, we propose a \textbf{R}isk-\textbf{A}ware \textbf{O}verlapping \textbf{S}ubgraph Learning framework (RAOS). 
We acknowledge that maintaining complete risk diffusion paths requires the replication of essential nodes. Crucially, risk patterns often propagate through long-tail users, which are vulnerable to graph partitioning. Therefore, we first initialize subgraphs to guarantee full node coverage with balanced computation. To restore the severed contexts, we introduce a budget-constrained overlapping sampling strategy that selectively replicates informative long-tail nodes while filtering out noise, thereby preserving critical risk diffusion patterns and avoiding redundancy. We apply a dynamic heterogeneous graph encoder to capture the evolving user-risk interactions within each subgraph and derive spatio-temporal representations. Furthermore, we design a cross-subgraph consistency alignment mechanism. We harmonize diverse local representations into a globally consistent latent space via alignment constraints on overlapping nodes. 
In the offline and online evaluation of a billion-scale industrial credit dataset from Weixin Pay, our RAOS significantly excels in identifying risk users, providing a scalable solution for billion-scale deep graph learning. Our contributions are summarized as follows:
\begin{itemize}
    \item To the best of our knowledge, this is the first work to bridge the gap between performance and industrial scalability for GNNs deployment in credit risk detection by proposing an overlapping subgraph learning framework designed for billion-scale real-world graphs.
    \item We integrate budget-constrained risk context reservation with consistency alignment, which selectively preserves informative long-tail risk patterns to ensure efficiency and aligns overlapping representations to maintain consistency.
    \item Extensive offline and online experiments demonstrate our model's superior performance in identifying risky users. It successfully empowers credit risk detection in Weixin Pay that safeguards the integrity of digital financial services and establishes a billion-scale deep graph learning paradigm.
\end{itemize}

\section{Preliminary}
\subsection{Background}
The ability to effectively detect credit fraud risk and distinguish between high-quality users and risky users is the cornerstone of credit security. Financial institutions can formulate differentiated risk control measures for users of different risk levels. Specifically, the objectives of credit risk detection include: detecting users exhibiting established risk patterns to facilitate targeted measures; and identifying latent risk users whose risks have not yet fully manifested to mitigate severe financial loss in advance.
Consequently, we leverage records of triggered risk events such as multi-platform fraud and suspicious cash-out to construct comprehensive user profiles. In this way, we can excavate explicit risk behaviors and map the correlations between users and specific risk events. 

To capture the complex interactions between users and risk events, we construct a dynamic heterogeneous graph. The graph is composed of two distinct types of nodes: \textit{user}, representing the massive user base, and \textit{risk event}, representing specific risk behaviors (such as abnormal account activities). An edge is established between a user and a risk event when the user triggers the specific risk. Crucially, edges are weighted to quantify risk level and are time-sensitive. Edges can be added, deleted, or updated over time. The model needs to capture the temporal evolution of risk.
While graph structure provides a granular depiction of risk diffusion, its deployment in an industrial setting faces a severe scalability bottleneck. The Weixin Pay credit graph comprises billions of nodes and edges with daily updates. Training the entire graph structure and features in a worker is computationally intractable. This necessitates a distributed GNNs learning paradigm. In this paper, we focus on paradigms based on pre-partitioned subgraphs for parallel computation as mini-batch strategies struggle with high cross-worker communication overhead and I/O latency~\cite{lin2023comprehensive}. 

\begin{figure*}
    \centering
    \includegraphics[width=1\linewidth]{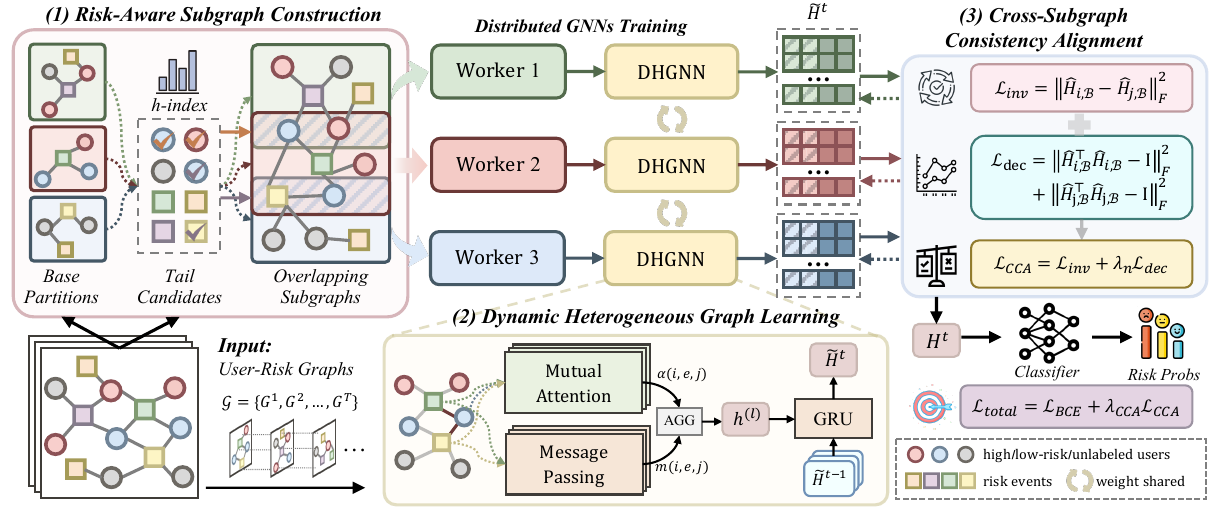}
    \caption{The overall architecture of the proposed RAOS framework. It proceeds in three stages: (1) Risk-Aware Subgraph Construction generates overlapping subgraphs by augmenting base partitions with long-tail risk nodes; (2) Distributed Learning utilizes a Dynamic Heterogeneous GNN to model user-risk evolution; (3) Cross-Subgraph Consistency Alignment harmonizes the representations of overlapping nodes, optimizing the model with a joint prediction and alignment objective.}
    \label{fig:model}
\end{figure*}

\subsection{Problem Formulation}
Formally, we define a sequence of dynamic user-risk heterogeneous graphs $\mathcal{G} = \{G^1, G^2, \dots, G^T\}$, where $T$ is the number of time steps. Each snapshot $G^t = (\mathcal{V}^t, \mathcal{E}^t)$ is a heterogeneous graph containing nodes and edges associated with specific types. In addition, each snapshot can be represented by its $d$-dimensional node attributes $X^t \in \mathbb{R}^{N \times d}$ and the corresponding adjacency matrix $A^t$, where $N$ is the number of nodes.
We define the mapping function $\tau(v): \mathcal{V}^t \rightarrow \mathcal{A}$ and $\phi(e): \mathcal{E}^t \rightarrow \mathcal{R}$ for each node $v \in \mathcal{V}^t$ and each edge $e \in \mathcal{E}^t$ respectively. In credit risk detection, the node type set $\mathcal{A} = \{\tau_u, \tau_r\}$ comprises two distinct entities: $\tau_u$ representing user nodes and $\tau_r$ representing risk event nodes. 
For an edge $e=(i, j)$ with weight $w_{ij}^t$ linked from a source node $i$ to a target node $j$, the meta relation~\cite{sun2011pathsim, sun2012mining} is formally denoted as a triplet $\langle \tau(i), \phi(e), \tau(j) \rangle$. $\phi(e)^{-1}$ represents the inverse of $\phi(e)$.
As time evolves, the sets of users and edges are non-static. We aim to learn a mapping function $f_\theta$ capable of capturing complex spatio-temporal dependencies. Specifically, given the graph snapshots and initial node attributes, $f_\theta$ predicts the credit risk score $\hat{y}^t$ for users:
\begin{equation} 
\hat{y}^t = f_\theta \left( G^t, X^t, H^{t-1} \right) 
\end{equation}
where $\hat{y}^t \in \mathbb{R}$ is the predicted risk probability. $H^{t-1}$ represents the latent historical representations.
The objective is to minimize the prediction error $\mathcal{L}(\hat{y}, y)$ on the labeled training nodes.

\section{Methodology}
To address the scalability challenges in industrial credit risk detection, we propose a Risk-Aware Overlapping Subgraph Learning framework. 
We adopt a distributed training paradigm based on the subgraphs that can be parallelized across distributed workers. 
As illustrated in Figure~\ref{fig:model}, the framework pipeline consists of three components:
(1) \textbf{Risk-Aware Subgraph Construction}: We first decompose the full graph into balanced base partitions to ensure computational load balance. Then we selectively expand the local subgraphs with informative long-tail nodes, ensuring that the subtle but critical risk diffusion paths are preserved within a single worker while structural noise is filtered.
(2) \textbf{Dynamic Heterogeneous Graph Learning}: for each subgraph, we employ Dynamic Heterogeneous GNNs to aggregate spatial-temporal information and evolve historical states, deriving comprehensive node representations.
(3) \textbf{Cross-Subgraph Consistency Alignment}: We introduce a cross-subgraph alignment module to address the representation inconsistency arising from overlaps. Leveraging the Canonical Correlation Analysis (CCA)~\cite{andrew2013deep} objective, we force the model to harmonize diverse local representations into a globally consistent latent space. 
The final risk probability is derived based on the aligned node representations.

\subsection{Risk-Aware Subgraph Construction}
For a billion-scale graph, training on the full adjacency matrix $A$ is computationally intractable. Therefore, we propose a risk-aware overlapping subgraph construction strategy for two key purposes: (1) it amplifies the risk propagation signals originating from long-tail entities, and (2) it preserves the structural integrity of risk diffusion paths via overlapping nodes, effectively mitigating the information loss induced by edge cuts. 

\textbf{Graph Partition.} To ensure computational load balance across $K$ distributed workers, we first decompose the entire graph into disjoint base partitions. Here, we employ a random hash partitioning strategy~\cite{gandhi2021p3} for its efficiency on billion-scale graphs. Note that the partitioning strategy is flexible to network characteristics and specific task requirements. 
Formally, the node set is divided into $K$ disjoint subgraphs $\mathcal{V} = [\mathcal{V}_1, \cdots, \mathcal{V}_K]$, and the feature matrix $X$ is divided into $[X_1, \cdots, X_K]$. Accordingly, the entire graph can be represented as a collection of subgraphs:
\begin{equation}
\bar{\mathcal{G}} = [G_1, \dots, G_K] = [\{\mathcal{V}_1, A_{1}, X_1\}, \dots, \{\mathcal{V}_K, A_{K}, X_K\}]
\end{equation}
The adjacency matrix $A$ is partitioned into $K^2$ submatrices as:
\begin{equation}
A = \begin{bmatrix} 
A_{11} & \cdots & A_{1K} \\ 
\vdots & \ddots & \vdots \\ 
A_{K1} & \cdots & A_{KK} \\ 
\end{bmatrix}
\end{equation}
where each diagonal block $A_{ii}$ represents the intra-partition links within $G_i$, which are preserved in the base partition. The off-diagonal blocks $\mathbf{A}_{ij}$ ($i \neq j$) contain the inter-partition links between $G_i$ and $G_j$.
In a disjoint distributed training setting, off-diagonal blocks $\mathbf{A}_{ij}$ are typically treated as zero matrices. 

\textbf{Budget-constrained Overlapping Sampling.} To address the issue of severed connections, we perform a budget-constrained sampling approach. We aim to selectively recover the critical structures in off-diagonal blocks $\mathbf{A}_{ij}$. For a subgraph $G_i$, we identify the structurally associated nodes $v \in \mathcal{V}_j (j \neq i)$ in other partitions. 
The associated nodes form the expansion candidate set $\mathcal{C}_i$:
\begin{equation}
\mathcal{C}_i = \{ v \in \mathcal{V} \setminus \mathcal{V}_i \mid \exists u \in \mathcal{V}_i, A_{vu} \neq 0 \}
\end{equation}

Practically, including all structural associates indiscriminately would introduce memory redundancy and irrelevant noise. Moreover, blindly selecting high-degree nodes can incorporate structural noise rather than valid risk signals. Instead, we hypothesize that critical risk signals often propagate through long-tail nodes~\cite{zhang2023deep, han2025mitigating} and they are vulnerable to partition-induced disconnection. 
We employ h-index~\cite{hirsch2005index} to quantify the structural influence of a node. We prioritize the tail nodes to augment the local subgraph. 
Specifically, we first filter candidates with lower h-index scores (h-index $\leq\theta$) and sample $b$ nodes to obtain the final expansion set $\mathcal{C}_{i,M}$, where $M$ is the predefined expansion budget that modulates the overlap ratio between subgraphs. An appropriate configuration ensures the preservation of non-redundant contexts while preventing structural noise. In this way, we selectively retain the informative long-tail contexts while filtering out noise. 
The final overlapping subgraph is constructed as the union of the base and expanded nodes: 
\begin{equation}
    \tilde{\mathcal{V}}_i = \mathcal{V}_i \cup \mathcal{C}_{i,M}
\end{equation}
It creates overlaps between subgraphs (i.e., $\tilde{\mathcal{V}}_i \cap \tilde{\mathcal{V}}_j \neq \emptyset$). 
Accordingly, we obtain an augmented adjacency matrix $\tilde{\mathbf{A}}_i$ that includes intra-subgraph links in $\mathbf{A}_{ii}$ and selected inter-subgraph links in $\mathbf{A}{ij}$. 
In this way, we can reserve risk diffusion paths via informative overlapping nodes. 
Based on augmented $\tilde{\mathcal{V}}_i$ and $\tilde{\mathbf{A}}_i$, we generate a sequence of graph snapshots $\{G_i^1, \dots, G_i^T\}$ by filtering nodes and edges based on timestamps, which serves as input for the subsequent graph learning module.

\subsection{Dynamic Heterogeneous Graph Learning}
For each subgraph, we propose a Dynamic Heterogeneous Graph Encoder (DHGNN) as the backbone $f_{\theta}$. It utilizes a Heterogeneous Graph Transformer (HGT)~\cite{hu2020heterogeneous} to capture spatial patterns, then incorporates an attention mechanism and GRU~\cite{dey2017gate} to model temporal dynamics. For simplicity, we omit the subscript $i$ of $G_i$ in this section and describe the learning process for a specific subgraph. 

\textbf{Heterogeneous Graph Transformer.} Let $h^{(l)}$ be the $l$-th latent representation at the current timestamp. Consider a target node $j$ and its neighbor $i \in \mathcal{N}(j)$ connected by an edge $e=(i, j)$. We parameterize the message passing process based on specific node and edge types. For a meta relation triplet $\langle \tau(i), \phi(e), \tau(j) \rangle$, we calculate the multi-head mutual attention score to estimate the importance of the source node $i$ to the target node $j$:
\begin{equation}
\text{Att-head}^s(i, e, j) = \left( K^s(i) W_{\phi(e)}^\text{Att} Q^s(j)^\top \right) \cdot \frac{\mu{\langle \tau(i), \phi(e), \tau(j) \rangle}}{\sqrt{d/S}}
\end{equation}
where $s$ denotes the index of attention heads, with $S$ being the total number of heads. $d$ represents the hidden dimension of the node representations. $K^s(i) = \text{Linear}_{\tau(i)}^s(h_i^{(l-1)})$ is the Key vector projected from the source node $i$'s feature. $Q^s(j) = \text{Linear}_{\tau(j)}^s(h_j^{(l-1)})$ is the Query vector projected from the target node $j$. $W_{\phi(e)}^\text{Att}$ is the trainable edge-based weight matrix. $\mu$ is a learned scalar prior that adaptively scales the attention weight for each meta relation.
Simultaneously, we compute the multi-head message by:
\begin{equation}
\text{Msg-head}^s(i, e, j) = \text{Linear}_{\tau(i)}^s(h_i^{(l-1)}) W_{\phi(e)}^{Msg}
\end{equation}
where $W_{\phi(e)}^{Msg}$ is the edge-based message projection matrix.
Finally, we normalize attention scores across neighbors and concatenate weighted messages. We get the updated vector through aggregation:
\begin{equation}
    \alpha(i, e, j) = \underset{i \in \mathcal{N}(j)}{\text{Softmax}} \left( \underset{s\in[1,S]}{\Vert} \text{Att-head}^s(i, e, j) \right)
\end{equation}
\begin{equation}
    m(i,e,j) = \underset{s\in[1,S]}{\Vert} \text{Msg-head}^s(i, e, j)
\end{equation}
\begin{equation}
    \tilde{h}_j^{(l)} = \underset{i \in \mathcal{N}(j)}{\oplus} \left( \alpha(i,e,j)\cdot m(i,e,j)
    \right) 
\end{equation}
where $\oplus$, $\Vert$ denote the aggregation and concatenation operation respectively. The final representation is updated via a linear projection followed by a non-linear activation and a residual connection:
\begin{equation}
    h_j^{(l)} = \sigma \left( \text{Linear}_{\tau(j)}^{Agg} \tilde{h}_j^{l} \right) + h_j^{(l-1)}
\end{equation}

\textbf{Temporal Graph Learning.} In dynamic user-risk graphs, node representations need to maintain structural proximity and capture temporal evolution~\cite{liu2026role}. Therefore, after obtaining the spatial embedding $h^t$, we integrate temporal information to generate a comprehensive risk probability of users. We incorporate a temporal attention mechanism and a GRU to learn informative historical states before performing the sequential update~\cite{liu2017global, cui2019hierarchical}. 
Let $C = [ H^{t-w} ; H^{t-w+1}; ... ; H^{t-1} ]$ be the context matrix consisting of recent $w$ hidden states. We normalize attention weights using a softmax function after a fully connected layer and $\tanh$ activation: 
\begin{equation}
a = \text{softmax} (r^{\top} \tanh \left(Q C^{\top}\right))
\end{equation}
where $r \in \mathbb{R}^{d}$, $Q \in \mathbb{R}^{d \times d}$ are the weight vector and matrix for attention, $d$ is the dimension of the hidden states. The weighted historical representation can be expressed as $\tilde{H}^{t-1} = (aC^{\top})$. 

We update the hidden states using the recurrence equation. Here, we adopt GRU and the final spatio-temporal node representation $H^t$ can be expressed as: 
\begin{equation}
H^t = \text{GRU} \left( h^t, \tilde{H}^{t-1}\right)
\end{equation}

\subsection{Cross-Subgraph Consistency Alignment}
As overlapping nodes are replicated across multiple subgraphs, a single user $u$ may possess distinct representations $H_{i,u}$ and $H_{j,u}$ derived from different subgraphs $G_i$ and $G_j$. These divergences arise because each subgraph aggregates a unique local neighborhood and temporal context. 
To resolve representation inconsistency, we introduce a cross-subgraph consistency alignment module. This module harmonizes local views into a globally consistent latent space.
Let $\mathcal{B} = \mathcal{V}_i \cap \mathcal{V}_j$ be the batch of overlapping nodes shared between subgraphs $G_i$ and $G_j$. The node embeddings obtained through the graph learning modules in the two subgraphs are denoted as $H_{i,\mathcal{B}} \in \mathbb{R}^{|\mathcal{B}| \times d}$ and $H_{j,\mathcal{B}} \in \mathbb{R}^{|\mathcal{B}| \times d}$. 
Directly minimizing the Euclidean distance between views often leads to dimensional collapse. To prevent this and eliminate redundancy in risk features, we adopt the node-level Canonical Correlation Analysis (CCA)~\cite{andrew2013deep, li2024hypergraph}. We aim to maximize the consistency between the $H_{i,\mathcal{B}}$ and $H_{j,\mathcal{B}}$
with the invariance term, while minimizing irrelevant information with the decorrelation term. 

First, we normalize the node embeddings along the batch dimension, ensuring that each feature dimension follows a distribution with a 0-mean and a $1/\sqrt{|\mathcal{B}|}$-standard deviation:
\begin{equation}\label{generateH}
    \hat{H} = \frac{H-\mu(H)}{\sigma(H) \times \sqrt{|\mathcal{B}|}}
\end{equation}
where $\mu(\cdot)$, $\sigma(\cdot)$ denote mean value and standard deviation for each feature dimension respectively. 
Next, we use the normalized $\hat{H}_{i,\mathcal{B}}$, $\hat{H}_{j,\mathcal{B}}$ to construct the CCA loss, which consists of an invariance term and a decorrelation term. The invariance term forces the representations of the same user from different subgraphs to be identical, ensuring the consistency of risk prediction. The decorrelation term minimizes the redundancy between different feature dimensions. 
\begin{equation} \label{eq:ccaloss1}
\mathcal{L}_{\text{inv}} =  \Vert \hat{H}_{i,\mathcal{B}} - \hat{H}_{j,\mathcal{B}} \Vert_F^2 
\end{equation}
\begin{equation}\label{eq:ccaloss2}
    \mathcal{L}_{\text{dec}} = \left\| \hat{H}_{i,\mathcal{B}}^\top \hat{H}_{i,\mathcal{B}} - I \right\|_F^2 + \left\| \hat{H}_{j,\mathcal{B}}^\top \hat{H}_{j,\mathcal{B}} - I \right\|_F^2
\end{equation}
\begin{equation}\label{eq:ccaloss3}
    \mathcal{L}_\text{CCA}= \mathcal{L}_\text{inv} + \lambda_n \mathcal{L}_\text{dec}
\end{equation}
where $I$ represents identity matrix, $\lambda_n$ denotes a non-negative weight, ${\Vert \cdot \Vert}_F^2$ indicates the squared Frobenius norm. 
Through this objective, the model is explicitly guided to solve representation inconsistencies caused by context divergence, yielding aligned embeddings across all subgraphs.

\subsection{Prediction and Optimization}
During the inference phase, to obtain the final risk probability for the entire graph, we aggregate the learned representations of all distributed workers. For overlapping nodes in multiple subgraphs, we compute their final embedding $H^t$ by averaging the aligned representations.  
Then, we can employ a Multi-Layer Perceptron (MLP)~\cite{rosenblatt1958perceptron} to predict credit risk. A two-layer MLP is expressed as:
\begin{equation}
\hat{y}^t = \text{MLP}(H^t) = W_2 \cdot \sigma(W_1 H^t + b_1) + b_2
\end{equation}
where $W_1$, $W_2$ and $b_1$, $b_2$ are trainable weights and biases. 

Instead of discarding unlabeled data, we adopt a semi-supervised learning paradigm~\cite{zhu2022leveraging}. The user attributes and risk signals are propagated among labeled and unlabeled users, which allows the model to capture hidden risk patterns from the structural context of neighbors. To achieve this, we filter the predictions to retain only those corresponding to the labeled users. Let $\hat{y}^t_m$ denote the masked predicted probabilities for the labeled subset, and $y^t_m$ denote the corresponding ground-truth labels.
We optimize the model by minimizing binary cross-entropy loss between the masked predictions and labels, which can be defined as:
\begin{equation}\label{eq:bceloss}
\mathcal{L}_{BCE} = - \frac{1}{N_m} \sum_{i=1}^{N_m} \left[ y_{m,i}^t \cdot \log(\hat{y}_{m,i}^t) + (1 - y_{m,i}^t) \cdot \log(1 - \hat{y}_{m,i}^t) \right]
\end{equation}
where $N_m$ is the number of labeled samples. 
Finally, the total objective function is defined as: 
\begin{equation} 
\mathcal{L}_{\text{total}} = \mathcal{L}_{BCE} + \lambda_{CCA} \mathcal{L}_\text{CCA}
\end{equation} 
where $\lambda_{CCA}$ is the hyper-parameter. The overall training process is summarized in Algorithm~\ref{alg:raos}. At each iteration, every worker performs forward propagation on its assigned subgraph in parallel and produces local gradients. Gradients are synchronized via AllReduce to update the shared model parameters. 
The model can be optimized through standard stochastic gradient descent-based methods. We use a default Adam optimizer~\cite{kingma2014adam} with the learning rate of $10^{-3}$ and the weight decay of $10^{-5}$ to execute the optimization process. 

\begin{algorithm}[t]
\caption{RAOS Training Algorithm}
\label{alg:raos}
\begin{algorithmic}[1]
\REQUIRE 
    Global Graph $\mathcal{G}=(\mathcal{V}, \mathcal{E}, X)$, 
    Labels $Y$, 
    Partitions $K$, 
    Budget $M$, 
    Threshold $\theta$, 
    Hyper-params $\lambda_{CCA}, \lambda_{n}$.
\ENSURE Optimized Model Parameters $\Theta^*$.
\STATE $\{\mathcal{V}_1, \dots, \mathcal{V}_K\} \leftarrow \text{Partition}(\mathcal{V}, K)$
\FOR{$k = 1$ to $K$ \textbf{in parallel}}
    \STATE $\mathcal{C}_k \leftarrow \{ v \in \mathcal{V} \setminus \mathcal{V}_k \mid \exists u \in \mathcal{V}_k, A_{vu} \neq 0 \}$
    \STATE $\mathcal{C}'_k \leftarrow \{ v \in \mathcal{C}_k \mid \text{h-index}(v) \leq \theta \}$
    \STATE $\mathcal{C}_{k,M} \leftarrow \text{TopM}(\mathcal{C}'_k)$
    \STATE $\tilde{\mathcal{V}}_k \leftarrow \mathcal{V}_k \cup \mathcal{C}_{k,M}; \quad \tilde{\mathcal{G}}_k \leftarrow \mathcal{G}[\tilde{\mathcal{V}}_k]$
\ENDFOR
\STATE Initialize parameters $\Theta$
\WHILE{not converged}
    \STATE $\mathcal{H}^t \leftarrow \emptyset$
    \FOR{$k = 1$ to $K$ \textbf{in parallel}}
        \STATE $H_k^t \leftarrow \text{DHGNN}(\tilde{\mathcal{G}}_k, X^t, \Theta)$ 
    \ENDFOR
    \STATE $H^t \leftarrow \text{MeanAgg}(\{ H_k^t \}_{k=1}^K)$
    \STATE $\hat{y} \leftarrow \text{MLP}(H^t, \Theta)$ 
    \STATE Compute $\mathcal{L}_{BCE}$ via Eq.(\ref{eq:bceloss})
    \STATE Compute $\mathcal{L}_{CCA}$ via Eq.(\ref{eq:ccaloss1} - \ref{eq:ccaloss3}) 
    \STATE $\mathcal{L}_{total} \leftarrow \mathcal{L}_{BCE} + \lambda_{CCA} \mathcal{L}_{CCA}$
    \STATE $\Theta^* \leftarrow \text{Adam}(\nabla_{\Theta} \mathcal{L}_{total})$
\ENDWHILE
\RETURN $\Theta^*$
\end{algorithmic}
\end{algorithm}

\begin{table*}[t]
\centering
\caption{Performance comparison on WeCreditFraud dataset across two snapshots. For each snapshot, we evaluate current ($T$) and forward ($T+90$) credit risk probability of users. The best and second best are highlighted in bold and underlined.}
\label{tab:main_result}
\begin{tabular}{l|l|cc|cc|cc|cc}
\toprule
\multirow{3}{*}{Category} & \multirow{3}{*}{Method} & \multicolumn{4}{c|}{$T_1$ (2024-12-31)} & \multicolumn{4}{c}{$T_2$ (2025-03-31)} \\ 
\cline{3-10} 
 &  & \multicolumn{2}{c|}{Current($T_1$)} & \multicolumn{2}{c|}{Forward ($T_1+90$)} & \multicolumn{2}{c|}{Current($T_1$)} & \multicolumn{2}{c}{Forward ($T_1+90$)} \\ 
 \cline{3-10} 
 &  & KS & AUC & KS& AUC & KS& AUC & KS & AUC\\ 
 \hline 
\multirow{3}{*}{\textit{Non-Graph}} 
 & XGBoost & $48.12{\scriptstyle \pm0.04 }$ & $80.85{\scriptstyle \pm0.02 }$ & $23.72{\scriptstyle \pm1.00 }$ & \bm{$66.14{\scriptstyle \pm0.56 }$} & 
 $44.87{\scriptstyle \pm0.04 }$ & \underline{$80.99{\scriptstyle \pm0.50 }$} &
 $21.90{\scriptstyle \pm1.00 }$ & 
 $65.04{\scriptstyle \pm0.56 }$ \\
 & MLP & $41.78{\scriptstyle \pm0.16 }$ & $78.30{\scriptstyle \pm0.28 }$ & $16.68{\scriptstyle \pm0.91 }$ & $61.00{\scriptstyle \pm0.16 }$ & $40.47{\scriptstyle \pm0.50 }$ & $77.32{\scriptstyle \pm0.15}$ & $17.57{\scriptstyle \pm0.05}$ & $60.72{\scriptstyle \pm0.36 }$ \\
 & PrefixSpan & $51.34{\scriptstyle \pm0.01 }$ & $75.67{\scriptstyle \pm0.03 }$ & $16.02{\scriptstyle \pm0.02 }$ & $58.01{\scriptstyle \pm0.01 }$ & $40.23{\scriptstyle \pm0.01 }$ & $70.12{\scriptstyle \pm0.01 }$ & $15.47{\scriptstyle \pm0.52 }$ &
 $57.73{\scriptstyle \pm0.26 }$ \\
 \hline
 \multirow{6}{*}{\textit{Graph}} & DHGNN & \underline{$51.96{\scriptstyle \pm0.03 }$} & \underline{$81.02{\scriptstyle \pm0.04}$} & \underline{$25.13{\scriptstyle \pm0.58 }$} & \underline{$65.93{\scriptstyle \pm0.42 }$} & \underline{$51.59{\scriptstyle \pm0.19 }$} & $80.58{\scriptstyle \pm0.02 }$ & \underline{$24.11{\scriptstyle \pm0.51}$} &  \underline{$65.17{\scriptstyle \pm0.31 }$}\\
 & \ -Random Hash & $42.46{\scriptstyle \pm0.78}$ & $76.51{\scriptstyle \pm0.14}$&
 $20.61{\scriptstyle \pm0.30 }$ & $63.23{\scriptstyle \pm0.21 }$ & $41.27{\scriptstyle \pm0.23 }$ & $75.93{\scriptstyle \pm0.08 }$ & $20.38{\scriptstyle \pm0.45 }$ &  $63.00{\scriptstyle \pm0.82 }$ \\
 & \ -Rule-based & $49.49{\scriptstyle \pm0.60 }$ & $78.41{\scriptstyle \pm0.23 }$ & $21.02{\scriptstyle \pm1.65 }$ & $62.49{\scriptstyle \pm1.07 }$ & $49.01{\scriptstyle \pm1.05 }$ & $78.72{\scriptstyle \pm0.34 }$ & $20.25{\scriptstyle \pm0.68} $ & $62.09{\scriptstyle \pm0.29 }$\\
 & \ -METIS & $51.73{\scriptstyle \pm1.01 }$ & $79.97{\scriptstyle \pm0.65 }$ & $22.18{\scriptstyle \pm0.07 }$ & $62.67{\scriptstyle \pm0.23 }$ & $50.67{\scriptstyle \pm0.07 }$ & $79.78{\scriptstyle \pm0.19 }$ & $22.30{\scriptstyle \pm0.70} $ & $63.41{\scriptstyle \pm0.04 }$\\
 & \ -COPRA & $52.60{\scriptstyle \pm0.97 }$ & $79.12{\scriptstyle \pm1.60 }$ & $21.51{\scriptstyle \pm1.90 }$ & $60.30{\scriptstyle \pm0.91 }$ & $51.46{\scriptstyle \pm1.27}$ & $78.72{\scriptstyle \pm2.15 }$ &  $21.84{\scriptstyle \pm0.29} $ & $61.11{\scriptstyle \pm1.22 }$\\ 
 \cline{2-10} 
 & \ -RAOS (Ours) & \bm{$52.96{\scriptstyle \pm0.06 }$} & 
 \bm{$81.57{\scriptstyle \pm0.01}$} & \bm{$26.47{\scriptstyle \pm0.16 }$} & $65.72{\scriptstyle \pm0.30 }$ & \bm{$52.27{\scriptstyle \pm0.38 }$} & \bm{$81.41{\scriptstyle \pm0.24 }$} & \bm{$25.43{\scriptstyle \pm0.68 }$} & \bm{$65.55{\scriptstyle \pm0.49 }$}\\ 
 \bottomrule
\end{tabular}
\end{table*}

\section{Experiments}
We conduct experiments to answer the following research questions: 
(1) \textbf{RQ1}: How is the overall performance of RAOS compared
with baselines on billion-scale credit risk detection? 
(2) \textbf{RQ2}: What is the contribution of various
components within RAOS to its overall performance?
(3) \textbf{RQ3}: How do hyper-parameter configurations affect the
performance of RAOS?
(4) \textbf{RQ4}: What real-world value does RAOS deliver when deployed in production?

\subsection{Experimental Setup}
\subsubsection{Datasets}
We conduct comprehensive experiments on a large-scale real-world dataset \textit{WeCreditFraud} provided by Tencent, sourced from the Weixin Pay platform. Data collection and utilization are strictly in compliance with security and privacy policies. The dataset spans a period from January 1, 2024, to March 31, 2025, and contains over 845M user behavior records involving approximately 498M unique users and covering over 60 distinct types of risk events. Based on the records, we construct dynamic user-risk heterogeneous graphs to picture correlations between users and diverse risk events. We derive a comprehensive risk score for each user by integrating multi-dimensional risk metrics, including severity and frequency of associated risk events. Notably, 88.3\% of users remain unlabeled due to insufficient credit records.

\subsubsection{Evaluations}
The primary objective of our framework is dynamic credit risk prediction. Given a sequence of $T$ graph snapshots, we aim to learn the spatio-temporal representations of users and output a scalar probability $\hat{y}_u^T \in [0, 1]$ at the final timestamp. 
To comprehensively assess the model's capability in both identifying existing fraudulent users and predicting latent risks, we conduct evaluations under two distinct settings based on different ground-truth label definitions: (1) \textbf{Current Risk ($T$)}: The labels are determined by the user's credit fraud records strictly at the observation date $T$. This setting evaluates the model's precision in detecting users who have already exhibited risky behaviors. (2) \textbf{Forward Risk ($T+90$)}: The labels are derived from the users' performance over a 90-day forward observation window $[T, T+90]$. Users are labeled as positive if they appear normal at $T$ but become fraudulent within the next 90 days according to the business definition. This setting tests the model's early-warning ability to uncover hidden risks before they materialize.

To comprehensively evaluate performance, we adopt two widely used metrics: KS (Kolmogorov-Smirnov)~\cite{massey1951kolmogorov} and AUC (Area Under the ROC Curve). The KS value quantifies the maximum distance between the empirical cumulative distribution functions of the predicted risk probability for users. A higher KS value indicates a superior capability to rank users based on risk. AUC evaluates the model's ability to discriminate between low-risk and high-risk users across various threshold settings.

\subsubsection{Baselines}
To comprehensively evaluate the effectiveness of our framework, we compare it with non-graph methods, including MLP~\cite{rosenblatt1958perceptron}, XGBoost~\cite{chen2016xgboost}, and PrefixSpan~\cite{han2001prefixspan}. For distributed GNNs training, we compared it with several graph partitioning algorithms, including Random Hash~\cite{gandhi2021p3}, Rule-based partitioning, METIS~\cite{karypis1997metis}, and COPRA~\cite{gregory2010finding}. More details can be seen in Appendix ~\ref{sec:baseline}.
Notably, we employ a unified dynamic heterogeneous graph backbone for graph-based comparisons. It ensures that any performance divergence is attributed to different subgraph partitioning strategies for distributed GNNs training. To establish an evaluation benchmark with complete topological information, we trained the DHGNN backbone on the entire graph structure via a mini-batch training paradigm with global neighbor sampling~\cite{hamilton2017inductive}. More detailed analysis with other backbones is reported in Appendix~\ref{sec:backbone}.

\subsubsection{Implementation Details}
For the backbone, we configure the Dynamic Heterogeneous Graph Neural Network with a hidden dimension of $d=64$. The HGT module utilizes 4 attention heads and $L=2$ layers to aggregate spatial information. For the temporal module, we set the window size $w=3$ to capture the evolution over the past three months. The overlap ratio is set to 0.3 and the h-index threshold is 2.
We set the non-negative weight $\lambda_n$ in $\mathcal{L}_\text{CCA}$ to 0.0002 and the contrastive loss weight $\lambda_{CCA}$ in the joint loss to 0.1.
We implement the distributed training framework on a server with 8 NVIDIA Tesla H20 GPUs. Our method is implemented using PyTorch 1.12.1 with CUDA 12.1 and Python 3.9 as the backend. The subgraphs are processed in parallel across the GPUs using PyTorch distributed data parallel framework.

\subsection{Credit Risk Detection Performance}
Table~\ref{tab:main_result} summarizes the performance on the WeCreditFraud dataset across two temporal snapshots ($T_1$ and $T_2$). We report the mean and standard deviation of KS and AUC metrics over five runs. Our proposed RAOS consistently achieves superior performance compared to all baselines for credit risk detection. Specifically, for immediate risk detection, RAOS significantly outperforms standard distributed strategies. Compared to Random Hash, which severs risk paths, RAOS achieves a KS improvement of over 10.5\%. This demonstrates that our overlapping strategy effectively restores the local structural context required to identify existing fraudsters. The advantage is even more pronounced in the forward risk assessment. While graph-based methods like COPRA achieve decent scores in current risk detection, their forward KS drops sharply. This indicates that COPRA's community-based expansion introduces excessive redundancy, which can mask subtle signals of latent risk and lead to overfitting on immediate risk patterns. 
In contrast, RAOS exhibits superior generalization capabilities. The model maintains a high forward KS by focusing on long-tail connectivity and successfully captures latent risk diffusion. 
RAOS identifies users who appear normal currently but are structurally connected to risk sources via subtle propagation paths, thus enabling effective early warning.

Remarkably, our framework outperforms the full-graph training backbone. Intuitively, the full-graph setting is expected to serve as the theoretical performance upper bound as it preserves complete topological information, whereas distributed methods inevitably incur information loss due to partitioning. However, real-world financial networks suffer from structural noise, where sparse risk signals are often overshadowed by redundant connections. We attribute our superior performance to the structural denoising effect of the long-tail overlapping strategy. By selectively sampling informative long-tail nodes while filtering out noise, our strategy enhances the representations of long-tail nodes. Consequently, this yields node embeddings that are more robust and discriminative compared to those derived from the noisy full graph.

\begin{table}
\setlength{\tabcolsep}{5pt}
\centering
\caption{Ablation study of RAOS on WeCreditFraud ($T_1$), comparing the overlapping strategy with varying sampling strategies and the consistency alignment module.}
\label{tab:ablation}
\begin{tabular}{l|cc|cc}
\toprule
\multirow{2}{*}{Variant} & \multicolumn{2}{c|}{Current ($T_1$)} & \multicolumn{2}{c}{Forward ($T_1+90$)} \\ 
\cline{2-5} 
 & KS & AUC & KS & AUC  \\ 
 \hline
Random Hash & 42.46 & 76.51 & 20.61 & 63.23  \\ 
+ Overlap (Random) & 51.57 & 80.65 & 24.35 & 65.01 \\
+ Overlap (high h-index) & 48.33 & 79.09 & 24.45 & 65.07 \\
+ Overlap (low h-index) & 52.90 & 81.27 & 25.14 & 65.09 \\ \hline
\ \ \  + Align (\textbf{=RAOS}) & \textbf{52.96} & \textbf{81.57} & \textbf{26.47} & \textbf{65.72}  \\ 
\bottomrule
\end{tabular}
\end{table}

\subsection{Ablation Study}
To verify the contribution of each component in our framework, we conduct an incremental ablation study on WeCreditFraud ($T_1$). We start with the Random Hash baseline and sequentially incorporate the overlapping strategy, varying overlapping sampling mechanisms, and consistency alignment. The results are summarized in Table~\ref{tab:ablation}, validating the following conclusions:
(1) The overlapping strategy is essential. We observe a substantial performance leap upon introducing the overlapping mechanism. This confirms that disjoint partitioning severely disrupts risk propagation. Even a random overlapping sampling strategy effectively recovers a significant portion of the lost structural context, yielding improvements of 9.11\% and 3.74\% in current and forward KS, respectively.
(2) Valid risk patterns are concealed in long-tail entities. We compare the three sampling strategies that sample random, high h-index, and low h-index nodes. The variant prioritizing high h-index nodes performs worse than the random strategy, with current AUC dropping by 1.56\%. This indicates that indiscriminate expansion of high-degree hubs introduces structural noise and redundancy. Conversely, prioritizing long-tail nodes preserves critical risk context and amplifies valid risk signals, thereby yielding optimal results.
(3) Representation consistency is critical. Comparing the model with and without the alignment module, we observe that the performance gains, especially the forward KS shows a 1.33\% increase. This demonstrates that the alignment module effectively projects local views into a globally consistent latent space, which is crucial for model generalization and forward risk prediction.

\begin{figure}
    \centering
    \includegraphics[width=1\linewidth]{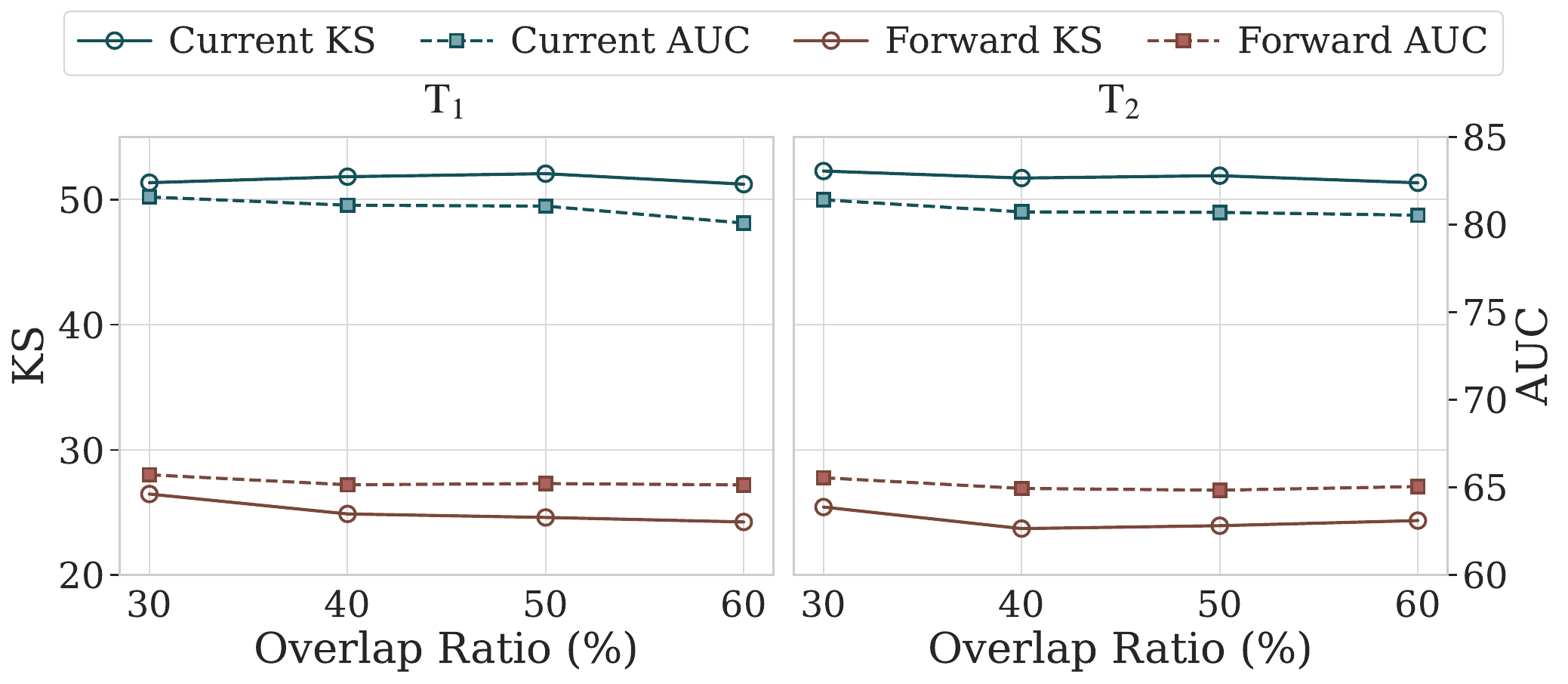}
    \caption{Impact of varying overlap ratio.}
    \label{fig:sensitive_overlapping_rate}
\end{figure}

\begin{figure}
    \centering
    \includegraphics[width=1\linewidth]{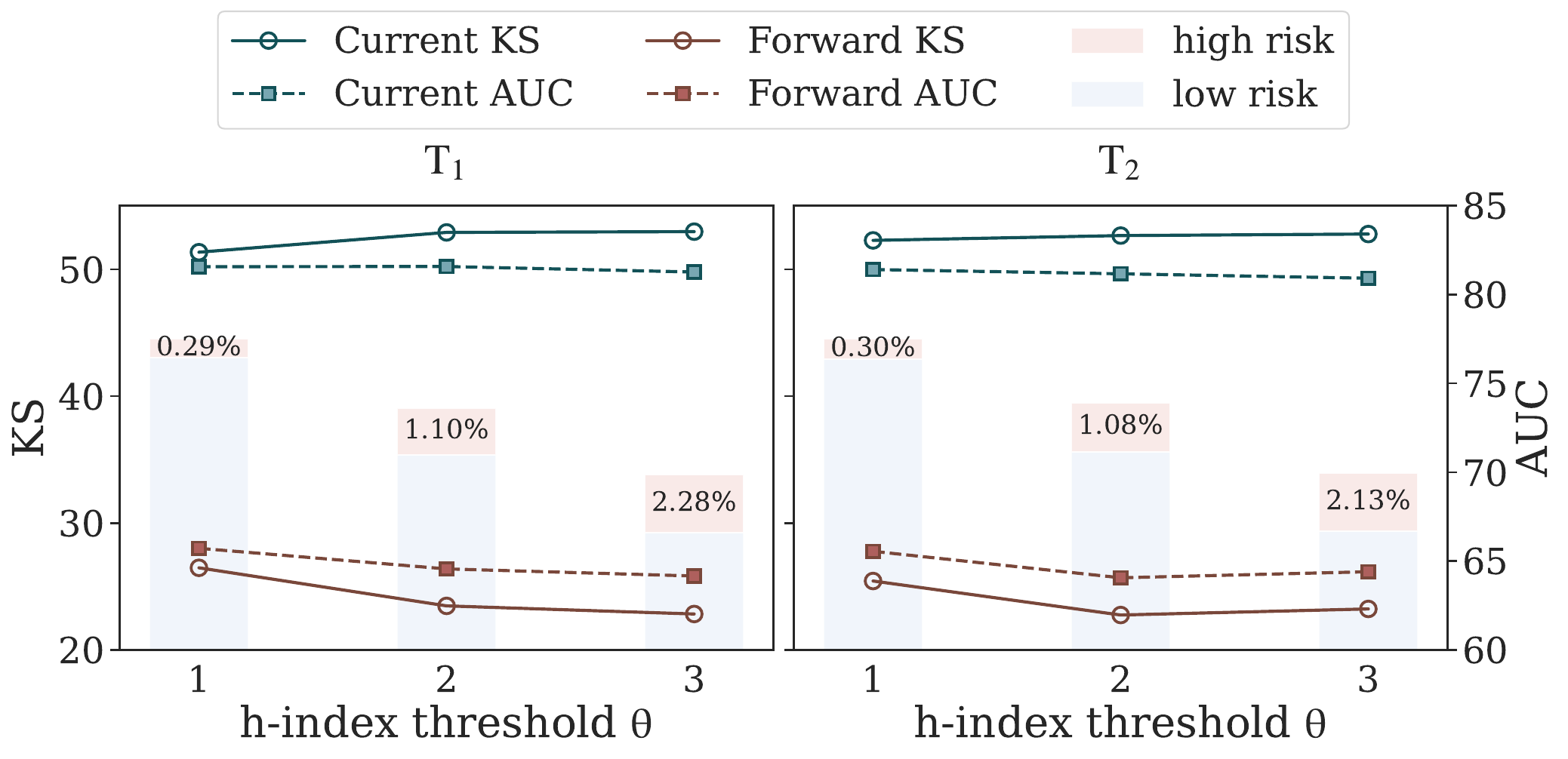}
    \caption{Impact of varying $\theta$. The line charts report the detection performance, while stacked bars show volume and risk composition of candidate nodes (h-index$\le\theta$). }
    \label{fig:sensitive_h-index}
\end{figure}

\subsection{Parameter Sensitivity Experiment}
We further discuss the important hyperparameters of varying overlap ratios and h-index thresholds. We conduct the parameter sensitivity analysis on WeCreditFraud ($T_1$ and $T_2$).

\textbf{Impact of varying overlap ratio.} The overlap ratio defined as $M/|V_i|$, represents the number of overlapping nodes $M$ relative to the size of the base subgraph $V_i$. 
As illustrated in Figure~\ref{fig:sensitive_overlapping_rate}, our RAOS exhibits a relatively stable performance as the ratio varies from 30\% to 50\%. It suggests that a controlled expansion budget is sufficient to effectively incorporate informative long-tail nodes and restore cross-subgraph connectivity, without introducing excessive redundancy. However, when the ratio keeps increasing, we observe a slight performance degradation. This indicates that an overly aggressive expansion strategy inevitably incorporates structural noise that dilutes the valid risk signals. Therefore, an appropriate selection of the overlap ratio represents an optimal trade-off, ensuring that the model enhances long-tail representation while maintaining a high signal-to-noise ratio. 

\textbf{Impact of varying h-index threshold.} 
The h-index threshold $\theta$ determines the selection scope for overlapping nodes. As illustrated in Figure~\ref{fig:sensitive_h-index}, the background stacked bars visualize the volume and risk composition of the candidate nodes satisfying h-index $\le \theta$. We observe a clear positive correlation between h-index and the risk ratio. As $\theta$ increases to include higher h-index nodes, the proportion of high-risk nodes rises to 2.28\%, validating the h-index as an effective indicator of risk density. 
However, this higher risk density does not translate into better generalization. While the current metrics show a marginal rise as the threshold increases, forward metrics suffer a notable decline. This suggests that high h-index nodes, despite being risk-dense, introduce excessive structural redundancy and noise that obscure subtle propagation paths. In contrast, low h-index nodes have sparse connectivity. They are more vulnerable to partitioning as the risk signals around them are easily extinguished. Therefore, limiting selection to low h-index nodes forces the model to focus on informative long-tail nodes. These nodes provide unique structural context that enhances robustness, preventing the model from overfitting to immediate high-risk signals and thereby preserving its capability to detect hidden risks.

\begin{figure}
    \centering
    \includegraphics[width=1\linewidth]{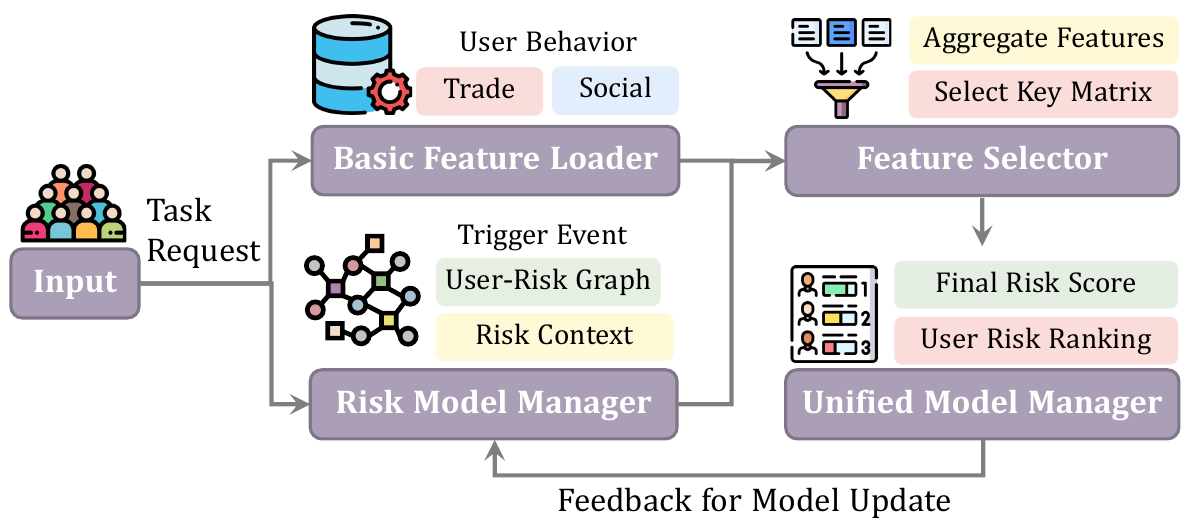}
    \caption{Overview of the online inference pipeline. Basic features and graph-based features are fused by a selector and utilized by a unified model manager to generate final risk scores and user rankings. A feedback loop continuously refines the model with online inference outcomes. }
    \label{fig:deployment}
\end{figure}

\section{System Deployment}
\subsection{Training and Inference}
\textbf{Distributed GNNs Training.}
Deploying a graph learning framework within Weixin Pay faces computational and storage challenges. We implemented a training system based on the distributed data parallel~\cite{li13pytorch} paradigm. Leveraging the overlapping subgraph strategy, we transform the global graph into self-contained training samples. This allows us to replicate the model across $K$ GPU workers and each worker performs forward propagation on its assigned subgraph in parallel. During back-propagation, gradients are synchronized globally using the Ring-AllReduce algorithm~\cite{gibiansky2017bringing}, ensuring linear scalability. In production, RAOS trains on WeCreditFraud on 8 H20 GPUs in 16 hours, with peak memory 66.6 GB per worker, communication and I/O overhead at 5.5\% and 0.75\% per epoch, and inference latency of 2.78 s for 1M users. These incremental costs remain within the production budget. More analysis about the communication overhead and training time scaling with the number of workers and overlap ratios is provided in Appendix~\ref{sec:scaling}. 
Regarding operational efficiency, we integrated the training workflow into Tencent’s tidal resource system. By leveraging the elastic GPU resource pool during off-peak hours, we optimize hardware utilization and significantly reduce deployment expenses without compromising model update frequency. 

\textbf{Online Inference.}
For the online serving phase, we designed an inference pipeline as illustrated in Figure~\ref{fig:deployment}. When a risk assessment task is triggered, a parallel processing mechanism is activated. On one track, the basic feature loader retrieves multidimensional behavioral data, including trade history, WeChat usage, and social interactions. Simultaneously, the risk model manager executes graph-based inference, leveraging the user-risk graph and overlapping subgraph structures derived from distributed training. These heterogeneous signals converge at the feature selector, where the system performs feature aggregation, core index counting, and key matrix selection to optimize the input representation. Subsequently, the unified model manager processes the fused features to generate both a final risk score and a user risk ranking for credit risk detection. The risk score and ranking are subsequently consumed by downstream risk-control modules. High-confidence risk users trigger immediate intervention measures, while medium-risk users enter a watchlist. Finally, the system employs a feedback loop from business experts, transmitting inference outcomes and subsequently observed ground-truth labels back to the training environment to continuously update and refine the models.

\begin{table}
\centering
\caption{Online A/B tests comparing with existing models on Weixin Pay platform in terms of User-level KS (User) and Amount-weighted KS (Amt) across different user segments.}
\label{tab:online_metrics}
\begin{tabular}{l|cc|cc|cc}
\toprule
\multirow{2}{*}{Model} & \multicolumn{2}{c|}{Overall} & \multicolumn{2}{c|}{Active} & \multicolumn{2}{c}{Inactive} \\ \cline{2-7} 
 & User & Amt & User & Amt & User & Amt \\ \hline
Baseline-v1 & 62.11 & 50.48 & 53.90 & 50.22 & 25.60 & 24.58 \\
Baseline-v2 & 62.86 & 52.37 & \textbf{55.59 }& 52.24 & 24.31 & 20.64 \\
\textbf{RAOS} & \textbf{63.16} & \textbf{52.70} & 55.09 & \textbf{52.52} & \textbf{28.37} & \textbf{26.05} \\ 
\bottomrule
\end{tabular}
\end{table}

\subsection{Online A/B Tests}
To thoroughly validate the real-world effectiveness and business impact of our framework, we deployed RAOS in the production environment of Weixin Pay and conducted a comparative analysis against existing incumbent models (referred to as Baseline-v1/v2). The A/B tests ran for 30 days with a 50/50 traffic split between the baselines and RAOS.
Unlike offline experiments, online evaluation requires a multidimensional perspective to balance user discrimination accuracy and financial asset safety. Therefore, we monitored two distinct indicators: \textit{user-level KS} and \textit{amount-weighted KS}. 
User-level KS measures the model's ability to distinguish between high-risk and normal users. It reflects the general discriminatory power over the population. Amount-weighted KS calculates the KS value based on the amount involved in each user's payment behavior. In financial credit risk control, a fraudulent case with a large amount causes significantly more damage than a small one. We partition the users into two complementary segments based on their payment activities. \textit{Active} users' frequent payment behaviors provide rich behavioral signals to the system, while \textit{Inactive} users' remain sparse, leaving existing models with limited individual evidence for risk assessment.
As presented in table~\ref{tab:online_metrics}, RAOS demonstrates comprehensive superiority. The most significant breakthrough is observed in the inactive segment. Compared with Baseline-v2, RAOS achieves a remarkable performance leap, increasing the user-level KS by 4.06\% and the amount-weighted KS by 5.41\% on inactive user risk detection. By preserving cross-subgraph risk context, our RAOS effectively captures the subtle risk signals of inactive users.

\begin{table}
\centering
\caption{Online A/B test results stratified by predicted risk probability. $\Delta$ denotes the relative improvement of RAOS over the baseline-v2 in risk user recall and risk amount recall.}
\label{tab:recall}
\begin{tabular}{l|c|c|c|c}
\toprule
Risk Probability & Top 1\% & (1\%-5\%] & (5\%-20\%] & (20\%, 100\%] \\ \hline
$\Delta$ User Recall   & +2.6\% & +0.7\% & +4.9\% & +3.9\% \\
$\Delta$ Amount Recall & +3.3\% & +3.5\% & +4.8\% & +7.5\% \\
\bottomrule
\end{tabular}
\end{table}

We further quantified the economic impact of RAOS on asset preservation. We compared both user recall and amount recall stratified by risk probability in table~\ref{tab:recall}. The two metrics measure the proportion of fraudulent users captured and the proportion of monetary loss prevented, respectively. We aligned the risk scores from both models and computed these recalls within the Top 1\%, 5\%, and 20\% risk score intervals. RAOS improves both metrics across all risk segments, with all gains being statistically significant ($p<0.01$). Specifically, the largest improvements concentrate in the Top (5\%-20\%] and (20\%, 100\%] segments, where risk user recall increases by 4.9\% and 3.9\% respectively compared to baseline-v2. The results validate RAOS's strength in mining concealed fraudulent users. Meanwhile, the gains in risk amount recall demonstrate that the framework's predictive precision effectively translates into tangible financial gains. These gains validate the model's ability to mine forward fraud risks, thereby preventing economic losses.

\section{Related Work}
\subsection{Credit Risk Detection}
Credit risk detection is necessary to maintain the stability of financial ecosystems and minimize financial loss. Existing risk detection methods have transitioned from rule-based systems to machine learning~\cite{cheng2018prediction, bulut2024comparison}. Methods like XGBoost have been widely adopted to asses users' financial risk by integrating detailed attributes~\cite{arora2020bolasso, aljadani2023mathematical}.
However, these methods heavily rely on feature engineering. Recently, graph-based methods have been proposed for risk detection to model the intricate topological structures of user-risk interactions \cite{li2024adarisk,xiang2025enhancing, guang2025homophily, cheng2023anti}. A heterogeneous graph comprising user and risk event nodes can be constructed to represent the risk landscape. GNNs can aggregate neighborhood information to uncover latent risk patterns. For example, HetGNN~\cite{zhang2019heterogeneous} samples heterogeneous neighbors via random walk. H2-FDetector~\cite{shi2022h2} uncovers hidden malicious entities by explicitly separating homophilic and heterophilic connections. Some researches have incorporated temporal dimensions to capture evolving user behaviors and sequential risk dependencies within dynamic graphs~\cite{cheng2020graph, cheng2020spatio}. However, the industrial deployment of GNNs faces a fundamental bottleneck: scalability. Since real-world graphs typically contain billions of nodes, they must be partitioned into subgraphs for distributed training. 

\subsection{Scalable Deep Graph Learning}
Scalability remains a critical bottleneck for deploying GNNs on billion-scale industrial graphs. Existing solutions primarily fall into two categories: (1) decoupling feature propagation from transformation to reduce computational complexity~\cite{lin2023comprehensive, wu2021seastar, zhou2023ugrapher}. SGC~\cite{wu2019simplifying}, HopGNN~\cite{chen2023node} simplifies GNNs via pre-computed propagation. NodeFormer~\cite{wu2022nodeformer} utilizes linearized attention mechanisms to bypass recursive message passing. 
In risk detection scenarios, such simplified mechanics often lack the adaptive sensitivity required to capture subtle non-linear risk patterns. 
(2) Sampling-based strategies have become the dominant paradigm for industrial applications~\cite{zeng2021decoupling}. GraphSAGE~\cite{hamilton2017inductive} samples a fixed number of neighbors per layer to constrain the receptive field, but suffers from the neighbor explosion problem. To mitigate this, subgraph-based methods including ClusterGCN~\cite{chiang2019cluster}, GraphSAINT~\cite{zeng2020graphsaint}, Ripple walks~\cite{bai2021ripple}, and learnable pooling strategies~\cite{gao2018large} decompose the graph into manageable subgraphs or grid-like structures for parallel training.
Those based on disjoint subgraphs often sacrifice structural integrity. This can be catastrophic for risk patterns characterized by subtle long-tail dependencies, as it indiscriminately severs the fragile evidence chains required for detection. 
While others consider overlapping subgraphs can restore risk contexts, the inclusion of overlapping nodes can introduce redundancy and noise. Furthermore, existing approaches lack a mechanism to align cross-subgraph representations of overlapping nodes, leading to semantic inconsistency. 


\section{Conclusion}
In this paper, we present RAOS, a scalable graph learning framework designed for billion-scale credit risk detection. To achieve both industrial scalability and high-precision detection, we devise a risk-aware overlapping subgraph learning framework. It effectively retains critical risk contexts, yet strictly controls the boundary expansion to mitigate noise and redundancy. Simultaneously, the framework can capture subtle risk patterns for long-tail users. Extensive offline evaluations and online deployment on billion-scale graphs demonstrate the effectiveness of RAOS as a scalable instrument for credit risk detection. It helps minimize financial losses and prevent the propagation of credit fraud risks, contributing to a stable financial ecosystem. Technically, our work can provide a reference for the industrial deployment of large-scale GNNs. 

\section{Ethics Statement}
This work strictly adheres to data privacy regulations and the internal data governance policies of Weixin Pay. All experiments were conducted on Weixin Pay's internal infrastructure. No raw user identifiers, message contents, or personally identifiable information were accessed by the authors at any stage of the pipeline. The graph encodes processed aggregated features (e.g., transfer counts between encrypted IDs), so that individual behaviors are neither exposed nor reconstructible.  

\begin{acks}
This work was supported by the National Science Foundation of China (Grant no. 62522213 and 62472317) and the Fundamental Research Funds for the Central Universities.
\end{acks}

\bibliographystyle{ACM-Reference-Format}
\bibliography{sample-base}

\appendix

\section{Detailed Descriptions of Baselines}
\label{sec:baseline}
To evaluate the proposed framework, we compare it against a diverse set of baselines, ranging from widely deployed industrial non-graph models to various scalable graph learning strategies. 

\textbf{Non-Graph Methods.} These methods treat users as independent and identically distributed instances, relying on statistical aggregations or sequential patterns. We feed them with aggregated user profiles and manual statistical features.
\begin{itemize}
\item \textbf{XGBoost}~\cite{chen2016xgboost}: An optimized distributed gradient boosting library that implements machine learning algorithms under the Gradient Boosting framework for efficient and accurate classification.
\item \textbf{MLP}~\cite{rosenblatt1958perceptron}: A standard feedforward neural network with fully connected layers, capable of learning non-linear features.
\item \textbf{PrefixSpan}~\cite{han2001prefixspan}: An efficient sequential pattern mining algorithm that explores prefix-projection to discover the complete set of frequent patterns.
\end{itemize}

\textbf{Graph-based Methods.} We evaluate RAOS against methods for scalable GNN training under our production setup. (1) We investigate the impact of different subgraph partitioning strategies on detection performance in distributed GNNs training. We employ a unified Dynamic Heterogeneous Graph Neural Network (DHGNN) backbone across all the following methods. The only variable is the partitioning algorithm used to distribute the graph across workers. (2) We compare with subgraph sampling training baselines adapted to distributed training. As the training cost of these paradigms exceeds 24 hours on the full production graph, we report their results on a sampled subgraph in the Appendix~\ref{sec:baseline}. 
\begin{itemize}
\item \textbf{Random Hash Partitioning}~\cite{gandhi2021p3}: A strategy that distributes nodes across partitions based on a hash function applied to node identifiers, ensuring uniform load balancing.
\item \textbf{Rule-based Partitioning}: A heuristic approach that groups nodes into partitions based on specific attribute values or predefined business logic. In our experiment, we partition users based on their business performance (e.g., transaction frequency, transaction volume, etc.)
\item \textbf{METIS}~\cite{karypis1997metis}: A multilevel strategy consisting of three phases: graph coarsening to reduce size, initial partitioning on the coarsest graph, and uncoarsening with refinement to project and optimize the partition on the original graph.
\item \textbf{COPRA}~\cite{gregory2010finding}: An efficient algorithm that extends label propagation to detect overlapping community structures in large networks by allowing nodes to propagate and retain membership in multiple communities simultaneously.
\item \textbf{GraphSAINT}~\cite{zeng2020graphsaint}: A subgraph-sampling paradigm where each worker independently samples a subgraph at every iteration via random walks, with per-edge normalization coefficients to correct estimator bias. 
\item \textbf{ClusterGCN}~\cite{chiang2019cluster}: A clustering-based paradigm that pre-partitions the graph into densely connected clusters using METIS, then forms training mini-batches by merging multiple clusters at each step. 
\end{itemize}

\begin{table*}
\centering
\caption{Comparisons on WeCreditFraud subgraph ($T_1$) in terms of detection accuracy and system metrics including per-epoch training time, I/O overhead, communication overhead, peak GPU memory, and inference latency.}
\label{tab:baseline}
\small
\begin{tabular}{l|cc|cc|rrrrr}
\toprule
\multirow{2}{*}{Method} & \multicolumn{2}{c|}{Current ($T_1$)} & \multicolumn{2}{c|}{Forward ($T_1+90$)} & \multirow{2}{*}{Train (s)} & \multirow{2}{*}{I/O (ms)} & \multirow{2}{*}{Comm (s)} & \multirow{2}{*}{GPU (GB)} & \multirow{2}{*}{Infer (ms)}\\ 
\cline{2-5} 
& KS & AUC & KS & AUC & & & & & \\
\midrule
GraphSAINT & 72.34 & 92.97 & 38.92 & 76.00 & 54.00 & 413.27 & 0.54 & 21.12 & 135.70 \\
ClusterGCN & 70.70 & 92.42 & 41.31 & 76.51 & 46.00 & 2.84 & 2.75 & 39.65 & 334.26 \\
DHGNN  & 72.29 & 92.87 & 42.08 & 76.74 & 4.78 & 0.01 & 0.01 & 39.42 & 223.25 \\
\ -Random Hash  & 72.70 & 92.44 & 40.24 & 76.58 & 3.92 & 1.29 & 0.28 & 20.43 & 242.08 \\
\ -RAOS (Ours) & \textbf{72.99} & \textbf{93.56} & \textbf{42.23} & \textbf{77.60} & 4.74  & 2.77 & 0.43 & 26.13 & 192.12 \\
\bottomrule
\end{tabular}
\end{table*}

\begin{table}
\setlength{\tabcolsep}{3.5pt}
\centering
\caption{Analysis of backbone generalization.}
\label{tab:backbone}
\begin{tabular}{c|l|cc|cc}
\toprule
\multirow{2}{*}{Backbone} & \multirow{2}{*}{Method} & \multicolumn{2}{c|}{Current ($T_1$)} & \multicolumn{2}{c}{Forward ($T_1+90$)} \\ 
\cline{3-6} 
 & & KS & AUC & KS & AUC  \\ 
 \hline
\multirow{2}{*}{GAT} & Random Hash & 74.21 & 93.87 & 37.78 & 74.19  \\
 & RAOS (Ours) & \textbf{75.22} & \textbf{94.31} & \textbf{38.55} & \textbf{75.86} \\ 
 \hline
\multirow{2}{*}{GraphSAGE} & Random Hash & 70.82 & 92.78 & 41.62 & 76.40 \\
 & RAOS (Ours) & \textbf{71.60} & \textbf{93.06} & \textbf{42.32} & \textbf{77.30} \\  
\bottomrule
\end{tabular}
\end{table}

\section{Additional Offline Experiments}
The supplementary analysis is performed on a representative subgraph by sampling at timestamp $T_1$ (2024-12-31), containing about 2M users and 30M edges.

\subsection{Comparison with Subgraph Sampling Baselines}
\label{sec:baseline}
We report detection accuracy together with system metrics, including training time per epoch, I/O overhead, communication overhead, peak GPU memory, and inference latency. As shown in table~\ref{tab:baseline}, RAOS consistently outperforms all baselines and remains efficient in system metrics. GraphSAINT suffers from accuracy degradation and a massive runtime I/O bottleneck because its online sampling tends to introduce redundant nodes and noise. ClusterGCN incurs a high communication cost for cross-worker cluster merging. In contrast, our RAOS avoids both bottlenecks via overlapping subgraphs, where cross-subgraph risk diffusion paths are preserved at a minimal runtime cost.

\subsection{Generalization to Different GNN Backbones}\label{sec:backbone}
While Table~\ref{tab:main_result} adopts DHGNN as the graph learning backbone, we further replace DHGNN with two widely used GNN backbones GAT~\cite{velivckovic2017graph} and GraphSAGE~\cite{hamilton2017inductive}, adapted to handle the heterogeneous user-risk graph. We compare disjoint random hash partitioning with RAOS partitioning under identical training settings. As shown in table~\ref{tab:backbone}, RAOS consistently improves all metrics across both backbones, yielding 1.01\%, 0.78\% improvements on current KS and 0.77\%, 0.70\% improvements on forward KS for GAT and GraphSAGE respectively. The results validate the effectiveness and generalizability of our proposed RAOS on diverse backbones. The effectiveness of RAOS is rooted in structure preservation rather than in coupling with a particular GNN architecture, supporting its applicability to a wide range of large-scale graph learning systems.

\begin{figure}
\centering
\subfloat[Communication overhead per epoch]{%
\includegraphics[width=0.5\linewidth]{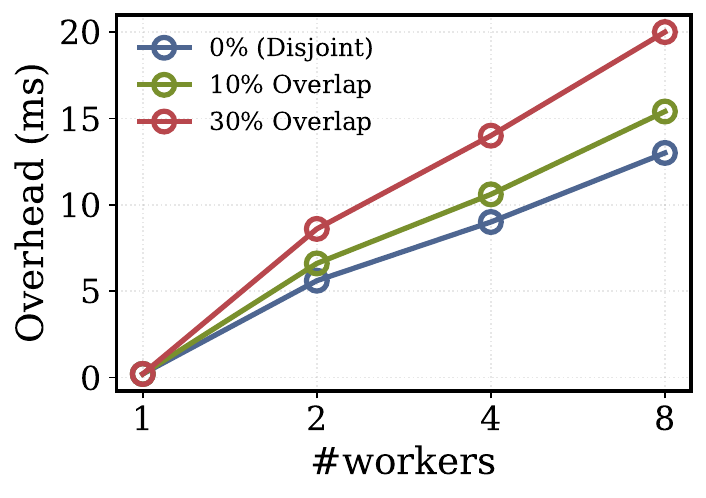}%
\label{fig:scaling_comm}}
\subfloat[Training time per epoch]{%
\includegraphics[width=0.5\linewidth]{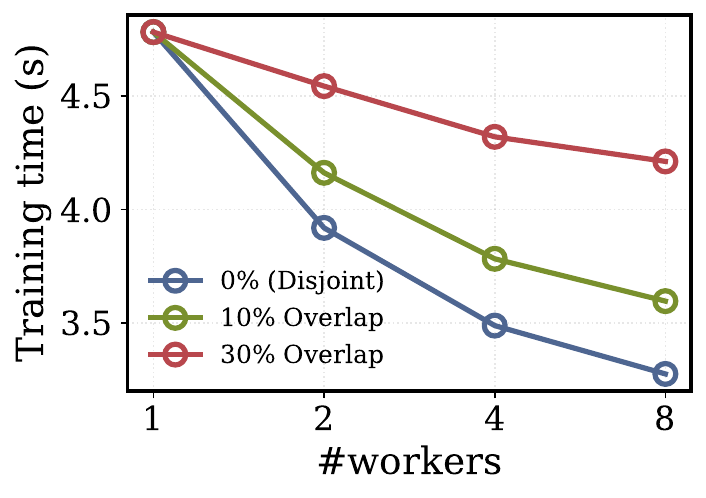}%
\label{fig:scaling_time}}
\caption{Scalability of RAOS across varying numbers of GPU workers and overlap ratios.}
\label{fig:scaling}
\end{figure}

\subsection{Communication Overhead Analysis}\label{sec:scaling}
We analyze the communication overhead of RAOS. Theoretically, the communication cost of RAOS arises from two sources: (1) gradient synchronization across workers via Ring-AllReduce and (2) representation alignment of overlapping nodes, which grows with the overlap ratio. 

Figure~\ref{fig:scaling} reports the communication overhead and total training time per epoch under varying numbers of GPU workers and overlap ratios. The communication overhead grows approximately linearly with the overlap ratio but remains a small fraction of the total training time. The training time per epoch of our RAOS scales near-linearly with the number of GPUs. Under disjoint partitioning, 8 workers achieve a 1.46$\times$ speedup over single-GPU training. The communication overhead increases with higher overlap ratios due to the additional cost of more overlapping nodes.

\end{document}